\documentclass[runningheads]{llncs}

\usepackage{eccv}

\usepackage{eccvabbrv}

\usepackage{graphicx}
\usepackage{booktabs}
\usepackage{array}
\usepackage{pifont}

\definecolor{mygray}{gray}{0.4}
\definecolor{darkgreen}{RGB}{0,128,0}
\newenvironment{mytiny}{\color{mygray}\small}{}
\usepackage[accsupp]{axessibility}  

\makeatletter\renewcommand\paragraph{\@startsection{paragraph}{4}{\z@}
	{.3em \@plus1ex \@minus.2ex}{-.5em}{\normalfont\normalsize\bfseries}}\makeatother

\usepackage[pagebackref,breaklinks,colorlinks,citecolor=eccvblue]{hyperref}

\usepackage{orcidlink}
\usepackage{float} 

\begin{document}

\title{DreamPolisher: Towards High-Quality Text-to-3D Generation via Geometric Diffusion} 

\titlerunning{DreamPolisher}

\author{Yuanze Lin\inst{1} \hspace{2mm}
Ronald Clark\inst{1} \hspace{2mm}
Philip Torr\inst{1}}

\authorrunning{Lin et al.}

\institute{$^1$ University of Oxford \\ 
\vspace{3mm}
\url{https://yuanze-lin.me/DreamPolisher_page/}
\vspace{-2mm}}

\maketitle

\begin{figure*}[!]
\centering
\advance\leftskip-0.1cm
\includegraphics[width=1.03\textwidth]{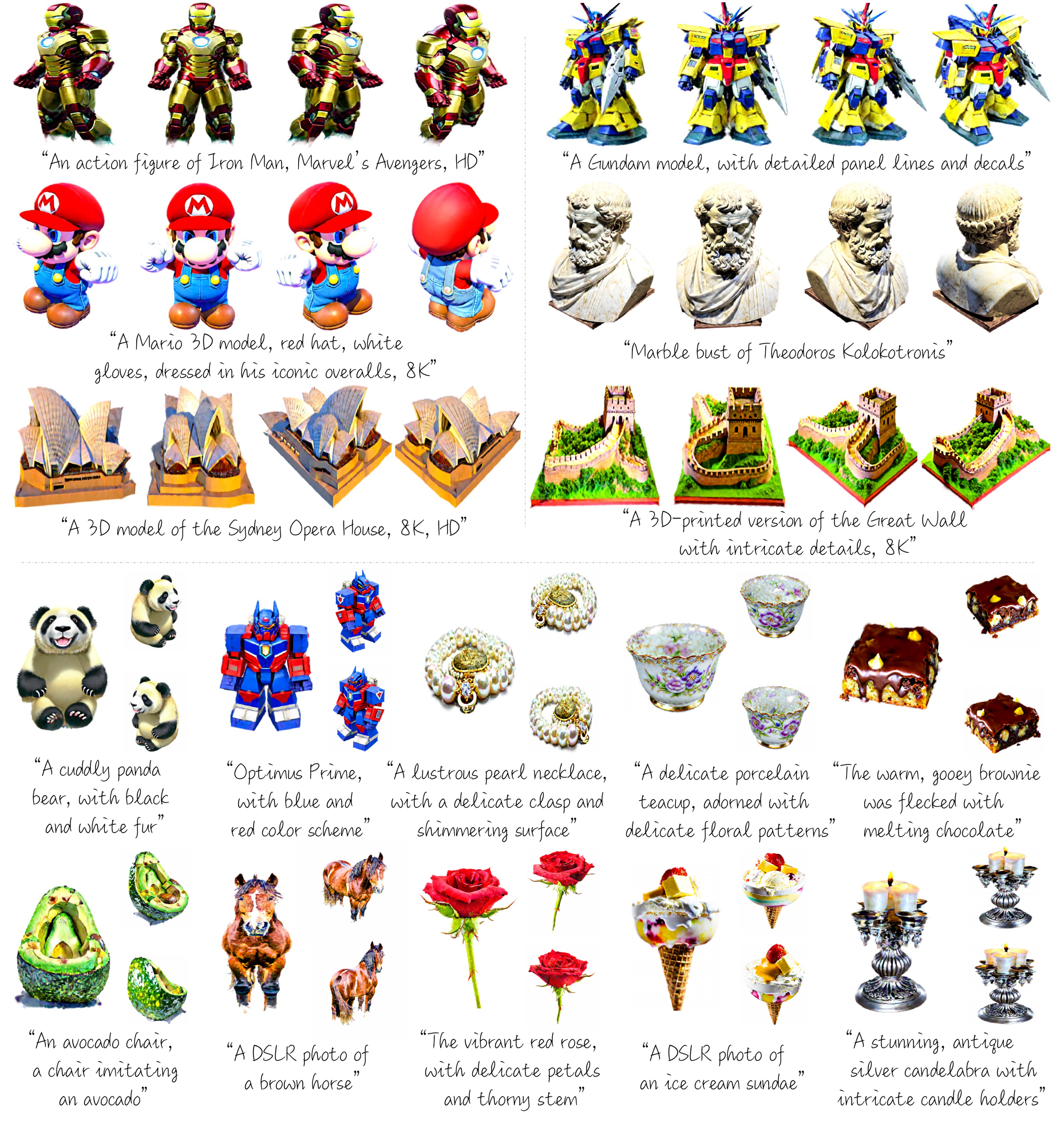}
    \caption{\textbf{Overview. } Given the user-provided textual instructions, DreamPolisher can generate high-quality and view-consistent 3D objects.}
\label{fig:figure1}
\end{figure*}

\begin{abstract}
  We present DreamPolisher, a novel Gaussian Splatting based method with geometric guidance, tailored to learn cross-view consistency and intricate detail from textual descriptions. While recent progress on text-to-3D generation methods have been promising, prevailing methods often fail to ensure view-consistency and textural richness. This problem becomes particularly noticeable for methods that work with text input alone. To address this, we propose a two-stage Gaussian Splatting based approach that enforces geometric consistency among views. Initially, a coarse 3D generation undergoes refinement via geometric optimization. Subsequently, we use a ControlNet-driven refiner coupled with the geometric consistency term to improve both texture fidelity and overall consistency of the generated 3D asset. Empirical evaluations across diverse textual prompts spanning various object categories demonstrate DreamPolisher's efficacy in generating consistent and realistic 3D objects, aligning closely with the semantics of the textual instructions. 
\end{abstract}
\section{Introduction}
In recent years, generative image and language models have demonstrated remarkable capabilities in generating high-quality content from textual input. Some prominent examples being text-to-text \cite{ouyang2022training,glaese2022improving,du2021glm,lin2022revive}, text-to-image \cite{kawar2023imagic,zhang2023adding,saharia2022photorealistic,ramesh2021zero,ramesh2022hierarchical,haque2023instruct,lin2023text,lin2023smaug} and text-to-video \cite{singer2022make,ho2022imagen}. The advances in text-to-image models \cite{rombach2022high,podell2023sdxl} have sparked great interest in using these models for 3D shape generation \cite{poole2022dreamfusion,lin2023magic3d}, which potentially opens up many possibilities for creating high-fidelity 3D content for applications such as virtual reality (VR), augmented reality (AR), gaming, movies, and architectural design.

Text-to-3D generation involves transforming user-provided textual descriptions into complex 3D models \cite{poole2022dreamfusion,lin2023magic3d,jain2022zero}. The predominant paradigm for text-to-3D focuses on training a parameterized 3D representation, \eg, NeRF \cite{mildenhall2021nerf}, to render high-quality images matching the given textual description by leveraging the strong guidance from a pretrained text-to-image diffusion model \cite{rombach2022high,podell2023sdxl}. The foundation of this capability lies in the use of Score Distillation Sampling (SDS) \cite{poole2022dreamfusion}. SDS has been shown to play a pivotal role in distilling 3D objects from 2D images generated by diffusion models, enabling the creation of high-quality 3D objects from simple text prompts.

Existing text-to-3D generation approaches can be categorized into two main classes, text-to-3D and text-to-image-to-3D. The former class \cite{tang2023dreamgaussian,yi2023gaussiandreamer,liang2023luciddreamer, wang2024prolificdreamer} mainly focuses on introducing new efficient 3D representations \cite{liang2023luciddreamer,tang2023dreamgaussian,yi2023gaussiandreamer}, such as 3D Gaussian Splatting \cite{kerbl20233d}, and modifying the SDS loss \cite{wang2024prolificdreamer,liang2023luciddreamer}, so that they can learn high-quality 3D shapes from the texture descriptions. However, such image-free methods usually lack details and are more susceptible to the ``Janus'' (\ie multi-face) problem \cite{shi2023mvdream}, compared to text-to-image-to-3D approaches.

On the other hand, text-to-image-to-3D methods \cite{sun2023dreamcraft3d,qian2023magic123} use a multi-view text-to-image diffusion model \cite{shi2023mvdream} to generate view-consistent images conditioned on text descriptions, and then introduce improved approaches\cite{shi2023mvdream,sun2023dreamcraft3d,qian2023magic123}, \eg, DMTet \cite{shen2021deep}, for refining the 3D parameterized models conditioned on both the images and textual description. This paradigm can achieve better quality for text-to-3D generation at the cost of longer training times. Based on the above observations, we seek to answer the question: ``can we bridge the quality gap between text-to-3D and text-to-image-to-3D methods?", in other words, we aim to generate \textit{view-consistent} and \textit{high-fidelity} 3D assets \textit{efficiently}.

To this end, we present DreamPolisher, a novel text-to-3D generation method based on 3D Gaussian Splatting. DreamPolisher first learns a coarse and view-consistent 3D objects initialized from an off-the-shelf text-to-point diffusion model and the textual description. It then learns to refine the obtained coarse 3D assets with a ControlNet refiner and geometric consistency loss, in order to significantly enhance the visual quality and view-consistency of the 3D objects. Extensive experiments with a diverse range of prompts containing various object categories show the superior performance of our proposed DreamPolisher approach compared with the existing state-of-the-art. 

Our contribution can be shortly summarized as the follows:

\begin{itemize}
\vspace{-2mm}
\item We present DreamPolisher, a novel text-to-3D generation approach, conditioned on text prompts and guided by geometric diffusion to obtain high-quality 3D objects.
\item Specifically, we introduce a ControlNet-based appearance refiner to further enhance the visual quality of the 3D assets, and also propose a new geometric consistency loss to ensure the view-consistency of 3D objects.
\item Experiments show the high quality and realism of our generated 3D assets, which demonstrates that our method outperforms existing text-to-3D generation methods.
\end{itemize}

\begin{figure*}[t!]
\centering
\includegraphics[width=1\linewidth]{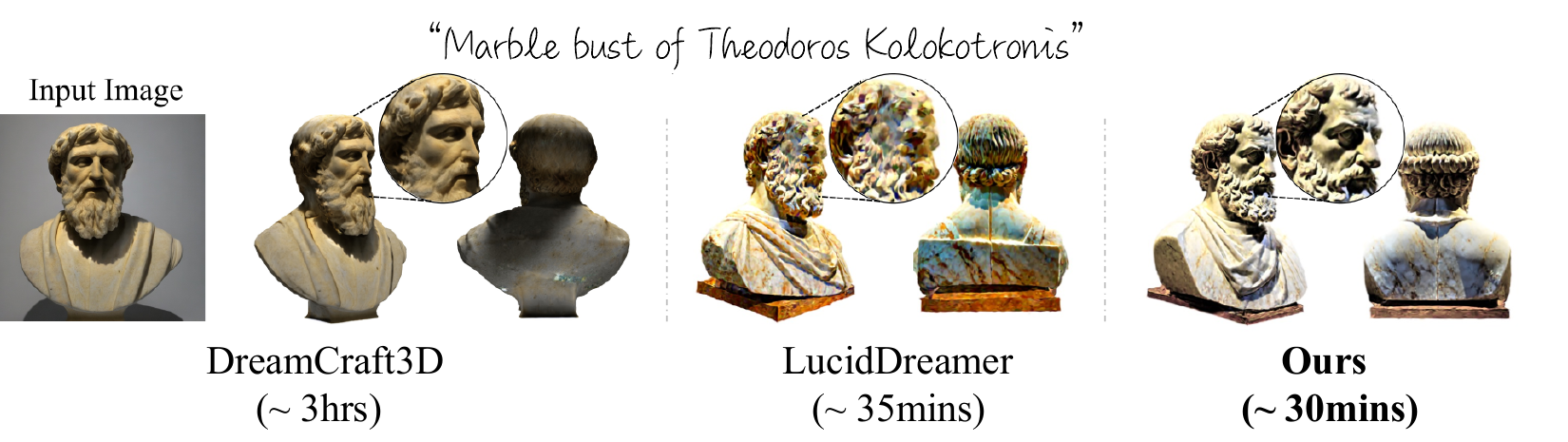}
\vspace{-3mm}
    \caption{\textbf{Comparison with existing methods.} The two predominant approaches for generating 3D objects from text are text-to-image-to-3D (\eg, DreamCraft3D \cite{sun2023dreamcraft3d}) and text-to-3D (\eg, LucidDreamer \cite{liang2023luciddreamer}) approaches. DreamCraft3D requires 3 hours and utilizes both a prompt and input image to generate a 3D object. While LucidDreamer can generate a 3D object with 35 minutes, it still struggles with consistency problems. Our method produces high-quality and visually coherent 3D objects quickly.}
\label{fig:figure2}
\end{figure*}

\section{Related Works}

\subsection{Text-to-3D Generation}
Recent advancements in 3D representation and diffusion models have significantly pushed the boundaries of text-to-3D generation. DreamField \cite{jain2022zero} optimized a Neural Radiance Field (NeRF) \cite{mildenhall2021nerf} across multiple camera views to produce rendered images under the guidance of a pre-trained CLIP model. Dreamfusion \cite{poole2022dreamfusion} introduced a novel loss formulation based on probability density distillation, specifically Score Distillation Sampling (SDS), enhancing the optimization of 3D models. This method enabled the generation of convincing 3D assets from text prompts by distilling existing text-to-image diffusion models, and sparked significant interest and a series of follow-up works \cite{shi2023mvdream, chen2023fantasia3d, lin2023magic3d, metzer2023latent, tang2023dreamgaussian, liang2023luciddreamer, tsalicoglou2023textmesh, chen2023text,yi2023gaussiandreamer, qian2023magic123}. Several SDS-based approaches \cite{chen2023fantasia3d, lin2023magic3d, metzer2023latent, sun2023dreamcraft3d, tang2023dreamgaussian} leveraged NeRF and other compact 3D representations \cite{tang2023dreamgaussian, yi2023gaussiandreamer} to significantly improve the visual fidelity of the generated 3D objects. 

Another line of research within text-to-3D generation seeks to refine the SDS formulation or introduce alternative score distillation methods \cite{wang2024prolificdreamer,yu2023text,katzir2023noise, liang2023luciddreamer}, with notable examples including ProlificDreamer \cite{wang2024prolificdreamer} and NSD \cite{katzir2023noise}, which employs variational score distillation (VSD) and Noise-Free Score Distillation (NFSD) to learn the parameters of the 3D representation, respectively.

A particular focus of some methods lies on ensuring cross-view consistency for 3D assets, addressing the so-called ``Janus'' problem. MVDream \cite{shi2023mvdream} fine-tunes the diffusion model with the multi-view images rendered from Objaverse \cite{deitke2023objaverse}. DreamGaussian \cite{tang2023dreamgaussian} and GSGEN \cite{chen2023text} integrated 3D Gaussian Splatting \cite{kerbl20233d} into the optimization of the 3D models via SDS. LucidDreamer \cite{liang2023luciddreamer} combined Gaussian Splatting with Interval Score Matching (ISM) to facilitate learning of the 3D objects. Our approach also utilizes Gaussian Splatting for 3D asset generation but improves on existing methods by incorporating ControlNet \cite{zhang2023adding} and geometric guidance to enhance the visual quality and achieve view consistency.

\subsection{Diffusion Models}
Diffusion models \cite{ho2020denoising, song2020denoising, nichol2021improved, rombach2022high,ramesh2022hierarchical,saharia2022photorealistic,podell2023sdxl} have emerged as a significant milestone in 3D generation \cite{shi2023mvdream,poole2022dreamfusion,qian2023magic123} and editing \cite{lin2023text, haque2023instruct, sella2023vox} tasks, notable for their exceptional quality of output. Denoising Diffusion Probabilistic Models (DDPM) \cite{ho2020denoising,saharia2022photorealistic,song2020score,rombach2022high} play an important role in text-driven image generation \cite{kawar2023imagic, saharia2022photorealistic} and editing \cite{lin2023text}. These models operate by using a Markov chain of diffusion steps that progressively add random noise to data and subsequently learn to predict the added noise in the reverse process, thereby reconstructing the original data samples. Stable Diffusion \cite{rombach2022high} introduced latent diffusion models, leading to significant enhancements in the training and sampling efficiency of denoising diffusion models. This approach enables text-aligned guidance for optimizing 3D models and has been extensively applied in recent text-to-3D generation approaches \cite{shi2023mvdream, liang2023luciddreamer, lin2023magic3d}.

Motivated by the previous text-to-3D approaches \cite{poole2022dreamfusion,shi2023mvdream,tang2023dreamgaussian,liang2023luciddreamer}, we also incorporate a text-to-image diffusion model (\eg, Stable Diffusion) as guidance to learn 3D model parameters.

\subsection{Differentiable 3D Representations}
The 3D representations play a pivotal role in reconstructing detailed 3D objects, which can render the images across different camera poses in a differentiable fashion. There are a large variety of different 3D representations \cite{mildenhall2021nerf,barron2022mip,li2023neuralangelo,chen2023mobilenerf,hedman2021baking, poole2022dreamfusion, wang2023score, wang2021neus, lin2023magic3d} that have been applied in various 3D-relevant tasks including reconstruction and generation. Neural Radiance Fields (NeRF) \cite{chen2023mobilenerf} proposes to render photorealistic novel views of scenes through optimizing a neural-network based scene representation, which has been adopted for text-to-3D generation \cite{lin2023magic3d,sun2023dreamcraft3d,poole2022dreamfusion} as well. However, it requires a lot of processing time due to the computationally expensive ray marching process. Several methods \cite{muller2022instant, fridovich2022plenoxels} have been proposed to accelerate the training time of NeRFs.

Recently, 3D Gaussian Splatting \cite{kerbl20233d} has shown  promise in achieving significant efficiency for 3D reconstruction. In this work, we propose DreamPolisher which uses 3D Gaussian Splatting as the 3D representation due to its remarkable efficiency.
\section{Preliminary}
In this section, we briefly explain the notation used throughout this paper.

\paragraph{Diffusion Models} \cite{ho2020denoising,rombach2022high} consists of forward diffusion and reverse diffusion processes. The forward diffusion process at timestep $t$ given the image $x_{t-1}$ with a scheduler $\beta_{t}$ can be defined as:

\begin{equation}
q(x_{t}|x_{t-1}) = \mathcal{N}(x_{t};\sqrt{1-\beta_{t}}x_{t-1},\beta_{t}\mathcal{I}).
\end{equation}

In the reverse diffusion process, the neural network $\theta$ is used to model the posterior $p_\theta(x_{t-1}|x_{t})$, which can be:
\begin{equation}
p_\theta(x_{t-1}|x_{t}) = \mathcal{N}(x_{t-1};\sqrt{\bar{a}_{t-1}}\mu_{\theta} (x_t), (1-\bar{a}_{t-1})\textstyle\sum\nolimits_{\theta}(x_{t})),
\end{equation}
in which $\mu_{\theta} (x_t)$ and $\textstyle\sum\nolimits_{\theta}(x_{t})$ represent the mean and variance of $x_{t}$, and $\bar{a}_{t-1} := (\prod_{1}^{t-1} 1-\beta_{t-1})$.

\paragraph{ControlNet} \cite{zhang2023adding} adds various spatial conditioning, \eg., depth, edges, segmentation, \etc, into a pre-trained text-to-image diffusion model \cite{ho2020denoising, song2020denoising} through ``zero convolutions" \cite{zhang2023adding}, so that the conditioning input can control the image generation. It utilizes the large-scale pretrained layers of diffusion models to construct a robust encoder. In our work, we leverage  ControlNet-tile to enhance the appearance details of the learned 3D objects. 

\paragraph{Interval Score Matching} \cite{liang2023luciddreamer} proposes a new objective to facilitate the training of text-to-3D generation. In particular, with the prompt $y$, the noisy latents $x_{s}$ and $x_{t}$ are obtained through DDIM inversion \cite{song2020denoising} from the rendered image $x_0$ of the 3D model. Here $x_0=g(\theta,\zeta)$, where $g(\cdot)$ and $\zeta$ refer to the differentiable renderer and camera pose, respectively. The ISM loss $\mathcal{L}_{\text{ISM}}$ over the 3D representation $\theta$ can be written as:

\begin{equation}
\mathcal{L}_{\text{ISM}}(\theta) := \mathbb{E}_{t,c}  \lbrack \omega(t) \lVert \epsilon_{\phi}(x_{t},t,y) - \epsilon_{\phi}(x_{s},s,\emptyset) \rVert^{2} \rbrack,
\end{equation}
where $\epsilon_{\phi}$ refers to the pre-trained diffusion model and $\omega(t)$ is the weight coefficient, $t$ and $s$ are two different timesteps. In this work, we adopt ISM instead of SDS due to its efficiency in generating high-quality 3D objects.

\begin{figure*}[t!]
\centering
\includegraphics[width=1\linewidth,height=4cm]{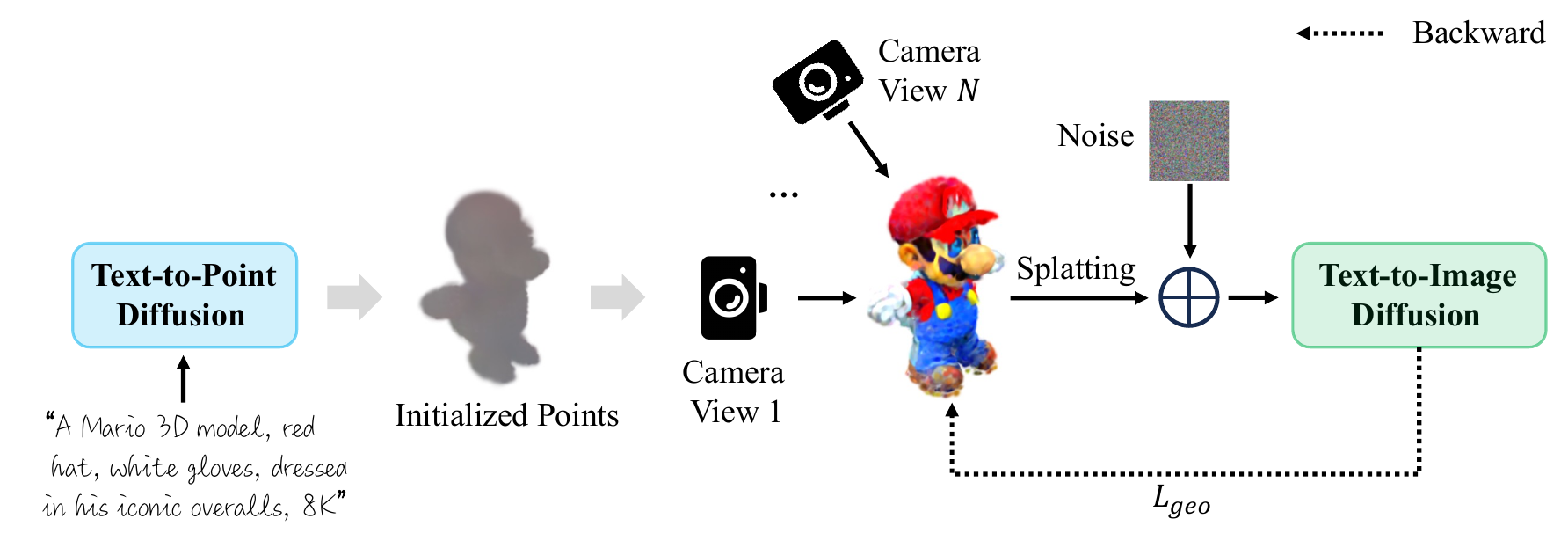}
\vspace{-7mm}
    \caption{\textbf{Coarse Optimization (Stage 1).} The text prompt is firstly fed into a pre-trained text-to-point diffusion model, \eg, Point-E\cite{nichol2022point} to obtain the corresponding point cloud, which is used to initialize the 3D Gaussians. After that, we use 3D Gaussian Splatting to optimize the object Gaussians guided by the pre-trained text-to-image diffusion model.
    }
\vspace{-3mm}
\label{fig:coarse_stage}
\end{figure*}

\section{Methodology}
To achieve our objective of generating 3D content with precise geometry and intricate textural details, DreamPolisher utilizes 3D Gaussians as the scene representation. Observing that a point cloud can be viewed as a collection of isotropic Gaussians, we initialize the Gaussians with a pre-trained point cloud diffusion model \cite{nichol2022point} to ensure consistent 3D geometry. This geometry prior helps mitigate the Janus problem and improves geometrical consistency. In the refinement stage, the Gaussians undergo iterative optimization with the geometry-guided ControlNet to enhance fine-grained details and coherence among views, while retaining the essential geometric information.

\subsection{Stage 1: Coarse Optimization}
\label{coarse_stage}
As shown in Figure \ref{fig:coarse_stage}, our method is built upon 3D Gaussian Splatting and enables the generation of text-aligned 3D objects. Given the text prompt $t$, we firstly feed it into a pre-trained text-to-point-cloud model, which can generate the initialized 3D points, after that, we optimize it and learn consistent 3D Gaussians for each object, guided by the text-to-image diffusion model \cite{rombach2022high}.

Since we adopt 3D Gaussian Splatting as the 3D representation, accurate initialization is essential for high-quality 3D asset generation. We firstly initialize the text-aligned 3D Gaussians by using the diffusion model, Point-E \cite{nichol2022point}, which can produce the 3D point cloud conditioned on a prompt. Building upon the initialized Gaussians by Point-E, we further optimize them through the ISM loss \cite{liang2023luciddreamer} to guide the coarse optimization, which is inspired by SDS and also uses a text-to-image diffusion prior. The gradient of the geometric loss $\mathcal{L}_{\text{geo}}$ over the 3D representation $\theta$ for the geometric object optimization can be written as: 

\begin{equation}
    \nabla_{\theta}\mathcal{L}_{\text{geo}} = {\mathbb{E}_{t,c}[\omega(t)(\epsilon_{\phi}(x_{t},t,y) - \epsilon_{\phi}(x_{s},s,\emptyset))\frac{\partial g(\theta,\zeta)}{\partial \theta}]}
\label{geo}
\end{equation}

To reconstruct coherent 3D objects with improved quality, we randomly render the images from the 3D objects with its corresponding camera pose, and optimize $\mathcal{L}_{\text{geo}}$ to better help generate consistent 3D shapes.

\begin{figure*}[t!]
\centering
\includegraphics[width=1\linewidth,height=4cm]{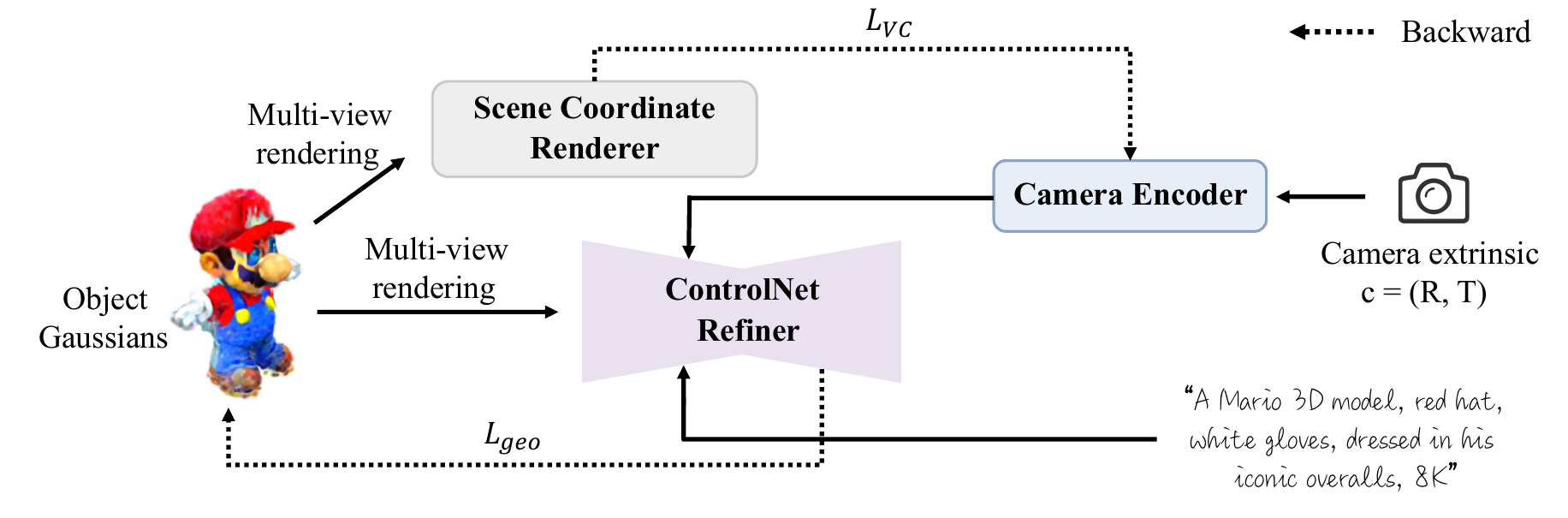}
\vspace{-8mm}
    \caption{\textbf{Appearance Refinement (Stage 2).} We render multiple views from the 3D object optimized by the coarse stage, and feed them to the Scene Coordinate Renderer. The rendered scene coordinates are then used in the view consistency loss, which aims to ensure that nearby scene points have consistent colors. The geometric embeddings from the camera encoder and the rendered multiple views are then fed into the ControlNet Refiner to generate high-quality and view-consistent 3D assets.} 
\vspace{-3mm}
\label{fig:refinement_stage}
\end{figure*}

\subsection{Stage 2: Appearance Refinement}
\label{refinement_stage}
Since the generated 3D objects from the coarse stage (\ie, Figure \ref{fig:coarse_stage}) typically lack details, we aim to further improve its quality. Thus, we introduce a refinement stage to enhance the appearance details. We show this refinement mechanism in Figure \ref{fig:refinement_stage}. We explain each component as follows.

\paragraph{Camera Encoder. } To ensure consistent refinement of details, such as colors and shades, across views, we condition the ControlNet refiner on the camera information. We specifically utilize the camera extrinsic matrix parameters \( \mathbf{T^c_w} = [\mathbf{R} | \mathbf{t}] \), where \( \mathbf{R} \in \mathbb{R}^{3\times3} \) is the rotation matrix, and \( \mathbf{t} \in \mathbb{R}^{3\times1} \) is the translation vector. These parameters are converted to the vector, $c \in \mathbb{R}^{1 \times 12}$, which is input into a camera encoder \( \psi \) to derive the camera-aware features:

\begin{equation}
f_{g}  = \psi (c).
\end{equation}

The camera-aware features $f_{g}\in \mathbb{R}^{1 \times d}$ are then injected into the ControlNet encoder's hidden states, conditioning the appearance refinement on camera directions.

\paragraph{ControlNet Refiner. } ControlNet \cite{zhang2023adding} is used for improving the textural details in the rendered views, and subsequently, in the generated 3D object.

We firstly sample multiple RGB views $\{v_{1}, ..., v_{N}\}$ from the coarse 3D object, $v \in \mathbb{R}^{C \times H \times W}$ and $N=4$ in our experiments. These sampled views are fed into a ControlNet-based Refiner, \ie, ControlNet-tile network $\varphi$, which can learn to provide more guidance to enhance the rendered views:

\begin{gather}
    \bar{v}_{i} = \varphi(v_{i}, \bar{f}, t) (i=1, ..., N), \\
    \bar{f} = [f_{t}, f_{g}],
\end{gather}
where $f_{t} \in \mathbb{R}^{m \times d}$ means the embedding of the textual instruction, and $\bar{v}_{i}$ denotes the refined RGB image which corresponds to view $v_{i}$.

\subsection{View-Consistent Geometric Guidance}
The refiner module we introduced in the previous section helps to improve textural quality, however issues such as multi-face inconsistency can still be an issue. Rather than relying solely on additional views for consistency, we introduce a novel geometric guidance, illustrated in Figure \ref{fig:refinement_stage}. We render two views of the object using the scene coordinate renderer to generate object-centric coordinate maps. These maps are then used to enforce visual and geometric consistency.

\paragraph{Scene Coordinate Renderer. } For a pixel coordinate \( \mathbf{u} = (u,v) \) in a sampled view, the scene coordinate renderer transforms \( \mathbf{u} \) into the corresponding world coordinate \( \mathbf{p_w} = (x_w, y_w, z_w) \). The transformation from world coordinates to pixel coordinates is:

\begin{equation}
    s \cdot \begin{bmatrix} u , v , 1 \end{bmatrix}^T =  \mathbf{K} \cdot \mathbf{T_w^c} \cdot \begin{bmatrix} x_w, y_w, z_w, 1 \end{bmatrix}^T,
\end{equation}
where \( s \) is a scaling factor, assumed to be 1 for simplicity, \( \mathbf{K} \) is the 3 $\times$ 3 intrinsic matrix and \( \mathbf{T_w^c} = [\mathbf{R} | \mathbf{t}] \) is the 3 $\times$ 4 extrinsic matrix that includes rotation \( \mathbf{R} \) and translation \( \mathbf{t} \). The transformation from pixel coordinate \( \mathbf{u} \) to camera coordinate \( \mathbf{p_c} \) can be written as:

\begin{equation}
    \mathbf{p_c} =  \mathbf{K}^{-1} \cdot \begin{bmatrix} u , v , 1 \end{bmatrix}^T,
\end{equation}
where \( \mathbf{p_c}=[x_c, y_c, z_c]^{T} \) represents the camera coordinate derived from the pixel coordinate. The camera-to-world transformation is given by \( \mathbf{T_c^w} = [\mathbf{R}^T | -\mathbf{R}^T \mathbf{t}]  \). Then, we can transform the camera coordinate \( \mathbf{p_c}\) to the world coordinate $\mathbf{p_w^{\prime}}$:

\begin{equation}
    \mathbf{p_w^{\prime}} =  \mathbf{T_c^w} \mathbf{p_c}.
\end{equation}

Normalizing the mapped world coordinate $\mathbf{p_w^{\prime}}$  involves converting the camera origin, $\mathbf{c_{0}}$, to world coordinates, and finding the ray from the camera center to the 3D point along the ray:

\vspace{-4mm}
\begin{align}
    \mathbf{c_{w}} &=  \mathbf{T_c^w} \mathbf{c_{0}}, \\
    \mathbf{e} &= \frac{\mathbf{p_w'}-\mathbf{c_{w}}}{\|\mathbf{p_w'}-\mathbf{c_{w}}\|}, \\
    \mathbf{p_w} &= \mathbf{c_{w}} + \hat{s} \cdot d \cdot \mathbf{e},
\end{align}
where \( \mathbf{c_{o}} = [0,0,0]^{T} \) represents the camera origin in the camera frame, \( \hat{s} \) as a scale factor set to 1, and \( d \) indicating the depth, or the distance between the view and the camera. This method maps all pixel coordinates of the rendered image \( I_{ij} \) to world coordinates \( \mathbf{p}^{ij}_w \). For notational simplicity we define a 3-channel, scene coordinate image, $S_{ij} = \mathbf{p}^{ij}_w$.

\paragraph{View-Consistency Loss. } 
Using the rendered scene coordinates, we introduce a novel view-consistency loss $\mathcal{L}_{\text{VC}}$, which enforces appearance consistency among views. The intuition behind this loss function is to force pixels that share a similar coordinate in the 3D scene space to assume similar color values.

Ideally, we would compare every pixel in the first image to every pixel in the second image by forming a Gram matrix, which would compare the similarity between all pixel pairs. However, this method faces a significant challenge due to computational cost. The size of the Gram matrix for an image would be $(H \times W)^2$ and would grow exponentially with the dimensions of the image, making it prohibitively expensive.

As a more feasible alternative, we choose to compare only overlapping pixels, which is inspired by the projective data association used in dense Iterative Closest Point (ICP) algorithms \cite{rusinkiewicz2002real}. This approach focuses on overlapping pixels that are projected from one view to another, assuming that these pixels, when they align spatially, should exhibit similar characteristics in terms of color and intensity. By limiting comparisons to these projectively associated pixels, we significantly reduce the computational burden while still effectively enforcing view consistency. The loss function is defined as:

\begin{gather}
    \mathcal{L}_{\text{VC}} = \frac{1}{2\times W \times H}\sum \nolimits_{i=1}^{H}\sum \nolimits_{j=1}^{W}(I_{ij} - \hat{I}_{ij})\cdot(1-\widetilde{S}_{ij}), \\
    \widetilde{S}_{ij} = S_{ij} - \hat{S}_{ij},
\end{gather}
in which, $\cdot$ means element-wise multiplication, $I_{ij}$ and $\hat{I}_{ij}$ are the RGB pixel values of two different sampled views, $S_{ij}$ and $\hat{S}_{ij}$ are the corresponding world coordinates of them, $W$ and $H$ are the width and height of the sampled view.

\paragraph{Training. } During the training stage, the view-consistency loss $\mathcal{L}_{\text{VC}}$ is used to optimize the camera encoder, so that it can learn the geometric embeddings which to be injected into ControlNet for cross-view coherence. We observe that using the view-consistency loss $\mathcal{L}_{\text{VC}}$ together with $\mathcal{L}_{\text{geo}}$ to directly optimize the 3D object model $\theta$ can lead to over-smoothing and be unstable, thus when optimizing $\theta$ in Stage 1, we only use $\mathcal{L}_{\text{geo}}$ for training.
\section{Experiments}

\subsection{Implementation Details}
\label{implementation}

\begin{figure*}[t]
\centering
\advance\leftskip-0.3cm
\includegraphics[width=1.11\textwidth]{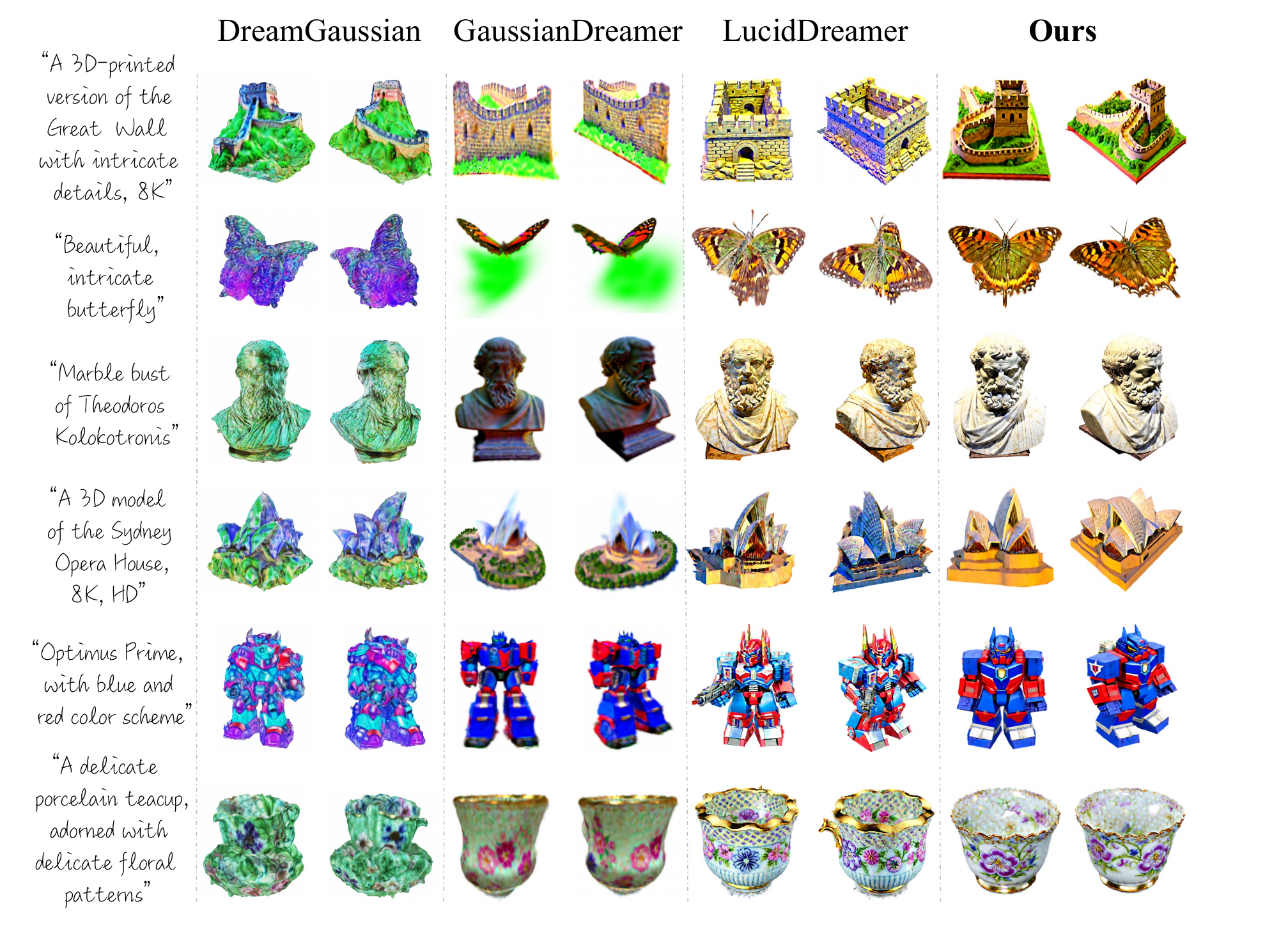}
\vspace{-6mm}
    \caption{\textbf{Comparison of existing text-to-3D approaches based on Gaussian splatting.} We compare the results with three state-of-the-art 3D Gaussian Splatting based approaches. The output generated by DreamPolisher achieves better consistency and has more intricate appearance details.} 
\vspace{-3mm}
\label{fig:fig6}
\end{figure*}

\paragraph{Guidance Model Setting. } We adopt a publically available text-image-diffusion model, Stable Diffusion \cite{rombach2022high}, for the coarse stage as the guidance model. In particular, we choose the Stable Diffusion 2.1 checkpoint\footnote{\url{https://huggingface.co/stabilityai/stable-diffusion-2-1}}. 
The view-dependent technique from Perg-Neg \cite{armandpour2023re} is used for training those asymmetrical objects. For the refinement stage, we use the ControlNet \cite{zhang2023adding} tile version, specifically ControlNet 1.1 tile\footnote{\url{https://huggingface.co/lllyasviel/control_v11f1e_sd15_tile}}. The guidance scale is 7.5 for all diffusion guidance models.
 
\paragraph{3D Gaussian Splatting Setting. } Our method is based on the 3D Gaussian Splitting (3DGS) pytorch CUDA extension\footnote{\url{https://github.com/graphdeco-inria/gaussian-splatting}}. To achieve better initialization for 3DGS, the pre-trained Point-E checkpoint \cite{nichol2022point} is used to generate the initial 3D point clouds. The Gaussian densification and pruning are performed from 100 iterations to 2500 iterations. For Gaussian densification, the Gaussians are split by the gradients every 100 iterations with a threshold $T_{\text{grad}}=0.00075$. For Gaussian pruning, the Gaussians are removed when their opacities are lower than 0.003. We progressively relax the view range every 500 training iterations. 

\paragraph{Training Details.} The training of DreamPolisher only requires 1 $\times$ 3090 GPU. We train the model 2500 and 1500 iterations during the coarse optimization and refinement stage respectively. For optimizing 3D Gaussians, the learning rates of the opacity, scaling and rotation are set to 0.05, 0.005 and 0.001. We adopt a learning rate of 0.001 for the camera encoder. The RGB images are rendered together with the corresponding depth maps from the 3D Gaussians for training.

\begin{figure*}[!t]
\centering
\advance\leftskip-0.35cm
\includegraphics[width=1.05\textwidth]{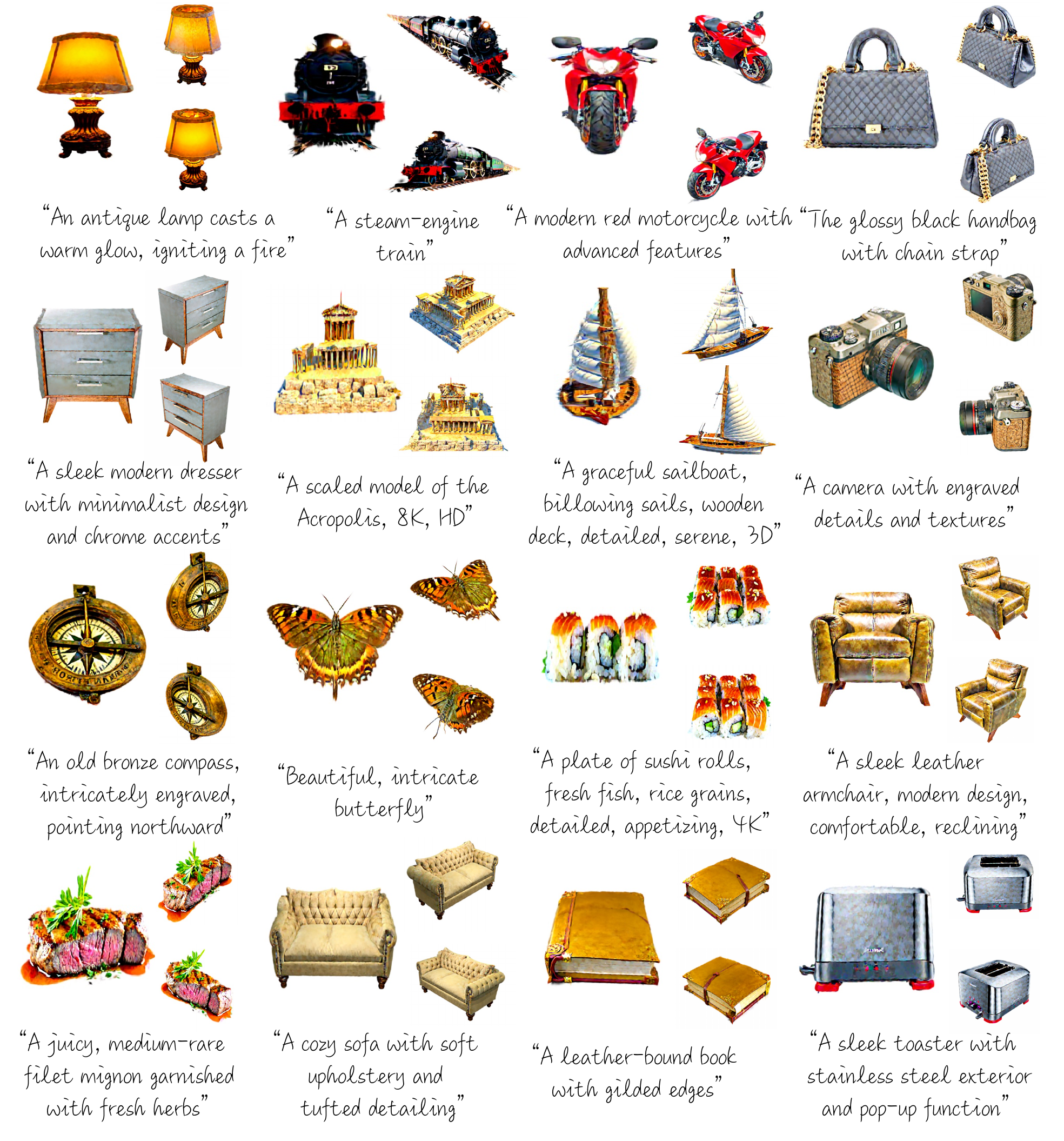}
\vspace{-6mm}
    \caption{\textbf{Generated text-to-3D examples across diverse object categories.} The proposed DreamPolisher shows strong generality by producing high-quality 3D objects conditioned on various textual prompts.}
\vspace{-1mm}
\label{fig:fig7}
\end{figure*}

\begin{figure*}[t]
\centering
\advance\leftskip-0.35cm
\includegraphics[width=1.05\textwidth]{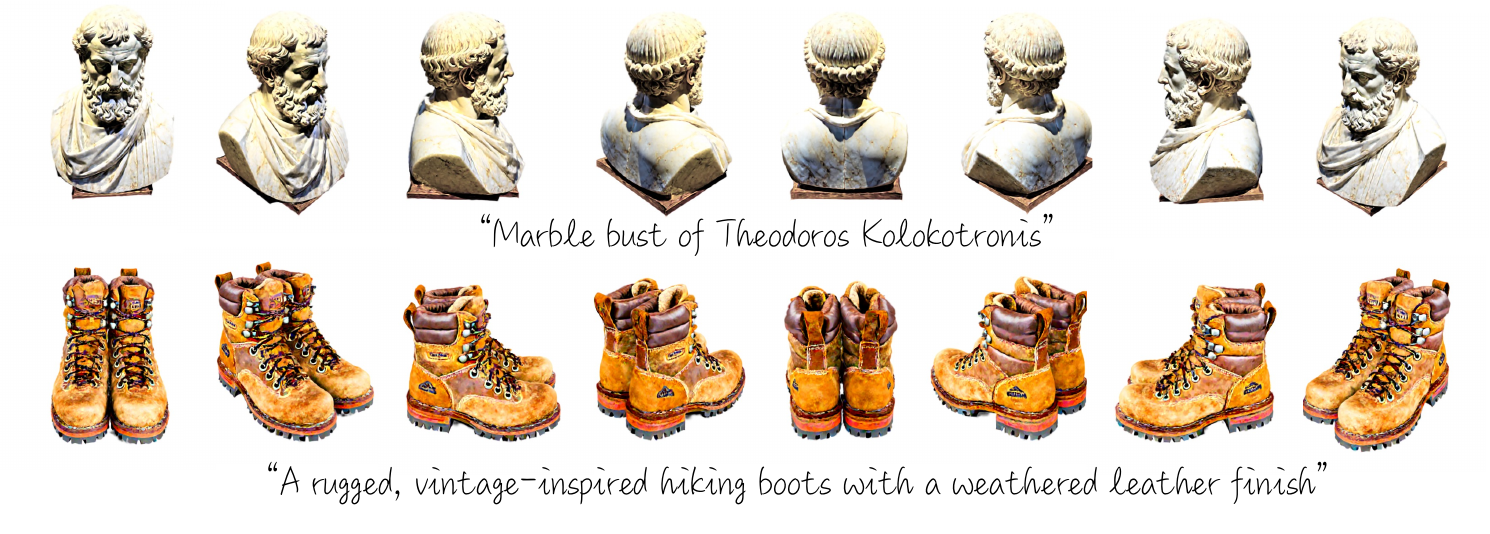}
\vspace{-8mm}
    \caption{\textbf{Different views of 3D objects generated by DreamPolisher.} The proposed DreamPolisher can generate view-consistent and high-fidelity 3D Gaussians given the textual instructions.}
\vspace{-2mm}
\label{fig:fig8}
\end{figure*}

\begin{figure*}[t]
\centering
\includegraphics[width=1.0\textwidth]{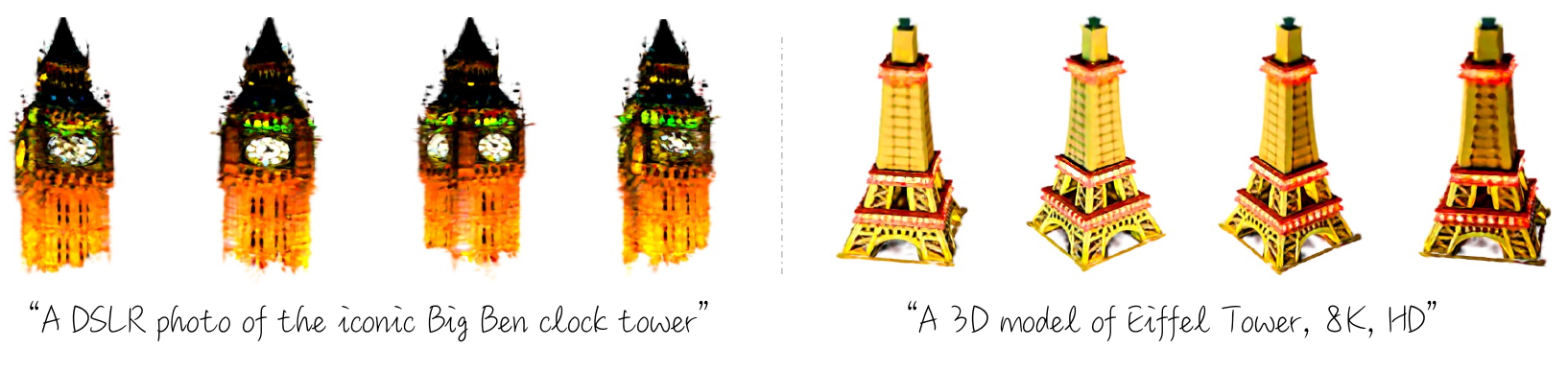}
\vspace{-8mm}
    \caption{\textbf{Failure cases.} We show two failure cases of our approach.} 
\vspace{-3mm}
\label{fig:fig10}
\end{figure*}

\begin{figure*}[t]
\centering
\vspace{-4mm}
\advance\leftskip-0.25cm
\includegraphics[width=1.03\textwidth]{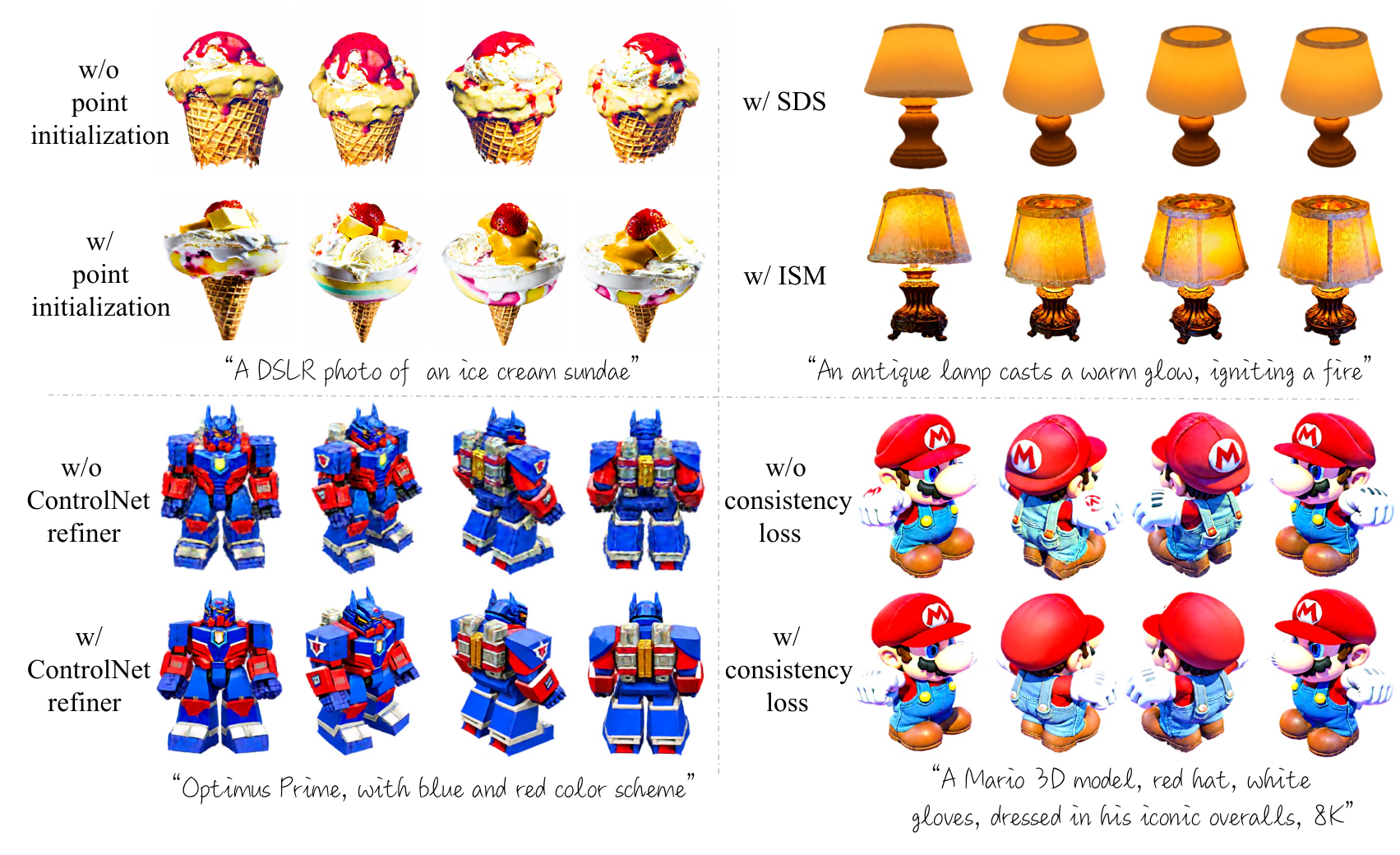}
\vspace{-6mm}
    \caption{\textbf{Results of the ablation study.} The left-top and right-top show the ablation studies of point initialization and the chosen SDS variant, while the left-bottom and right-bottom show the experiments of the ControlNet refiner and consistency loss.} 
\vspace{-2mm}
\label{fig:fig9}
\end{figure*}

\begin{table}[t]
\centering
\begin{tabular}{cp{3cm}<{\centering}cp{3cm}<{\centering}cp{3cm}<{\centering}cp{3cm}<{\centering}}
\toprule
Compared Methods & ViT-L/14 ($\uparrow$) & ViT-bigG-14 ($\uparrow$) & Generation Time \\ \midrule
Shap-E \cite{jun2023shap} & 20.51 & 32.21 & \textbf{6 seconds}\\
DreamFusion \cite{poole2022dreamfusion} & 23.60 & 37.46 & 1.5 hours\\
ProlificDreamer \cite{wang2024prolificdreamer} & 27.39 & \textbf{42.98} & 10 hours\\
Instant3D \cite{muller2022instant} & 26.87 & 41.77 & 20 seconds\\
GaussianDreamer \cite{yi2023gaussiandreamer} & 27.23  $\pm$\mytiny{0.06} & 41.88 $\pm$\mytiny{0.04} & 15 minutes\\
Ours & \textbf{27.35 $\pm$\mytiny{0.04}}  & 42.61 $\pm$\mytiny{0.05}  &  30 minutes \\ \bottomrule
\end{tabular}
\vspace{2mm}
\caption{\textbf{Quantitative results.} We show the results of quantitative comparisons of our approach to existing methods.}
\vspace{-9mm}
\label{tab:quantitative}
\end{table}

\subsection{Qualitative Comparison}
\label{comparison}
We compare our method to three state-of-the-art text-to-3D approaches: \textbf{(1) DreamGaussian} \cite{tang2023dreamgaussian} Uses a generative 3D Gaussian Splatting model \cite{kerbl20233d} combined with mesh extraction and texture refinement in UV space for improving details; \textbf{(2) GaussianDreamer} \cite{yi2023gaussiandreamer} bridges 3D and 2D diffusion models through 3D Gaussian Splatting, benefiting from both 3D coherence and intricate generation of details; \textbf{(3) LucidDreamer} \cite{liang2023luciddreamer} introduces Interval Score Matching (ISM), employing deterministic diffusing trajectories and leveraging interval-based score matching to mitigate over-smoothing, and uses 3D Gaussian Splatting for the 3D representation.  In all experiments, we use the official code from the compared approaches.

The comparisons results are shown in Figure \ref{fig:fig6}. We can observe that our method significantly surpasses existing methods, in terms of both the visual quality and consistency. 

\paragraph{Generality. } We show more results in Figure \ref{fig:fig7}, these objects are generated from diverse prompts encompassing categories such as food, vehicles, furniture, antiques, \etc. The generated objects show intricate details that correspond to the input prompts, showcasing the robust generality of our proposed method.

\paragraph{Consistency. } In Figure \ref{fig:fig8}, we show diverse views for different examples, there are 8 views for each object and these views represent evenly spaced azimuth angles from -180 to 180 degrees. We can see that the outputs produced by DreamPolisher consistently align across vastly different azimuth angles.

\paragraph{Failure Cases. } We show some failure cases from our method in Figure \ref{fig:fig10} to conduct a thorough analysis on the underlying causes. As depicted in Figure \ref{fig:fig10}, these failures may stem from inadequate supervision, particularly when the model relies solely on textual inputs. This limitation becomes more pronounced for objects that exhibit high symmetry and fine details.

\subsection{Quantitative Comparison}
We present quantitative comparisons using the CLIP similarity metric \cite{radford2021learning} with other methods in Table \ref{tab:quantitative}. Results for other methods are sourced from the concurrent GaussianDreamer paper \cite{yi2023gaussiandreamer}. The results for DreamFusion and ProlificDreamer are derived from the implementation by ThreeStudio\footnote{\url{https://github.com/threestudio-project/threestudio}}.

Following GaussianDreamer, we employ a camera radius of 4, an elevation of 15 degrees, and select 120 evenly spaced azimuth angles ranging from -180 to 180 degrees, resulting in 120 rendered images from distinct viewpoints. Subsequently, we randomly choose 10 images from the 120 rendered ones. Finally, the similarity between each selected image and the corresponding text is computed, we average the scores of these 10 examples as the final similarity. ViT-L/14 \cite{radford2021learning} and ViTbigG-14 \cite{schuhmann2022laion, ilharco_gabriel_2021_5143773} are used to calculate CLIP similarity. The performance of our method is close to ProlificDreamer and significantly surpass other methods. Furthermore, it only requires \textbf{5\%} of the generation time compared to ProlificDreamer.

\subsection{Ablation Study}
\label{ablation}
\paragraph{Initialization. } As shown in Figure \ref{fig:fig9}, employing the pre-trained text-to-point diffusion model \cite{nichol2022point} for point initialization results in significantly improved geometric shapes for the generated 3D objects, aligning more closely with the provided textual prompt.

\paragraph{The Effect of SDS Loss Variant. } Given that we choose the ISM loss to refine the 3D object's representation, we investigate the impact of this choice by conducting an ablation study comparing the use of SDS and ISM as the geometric loss $\mathcal{L}_{geo}$ for optimization throughout the entire training process. As illustrated in Figure \ref{fig:fig9}, it is evident that our approach yields 3D objects with significantly enhanced details when leveraging the ISM loss for training.

\paragraph{The Influence of the ControlNet Refiner. } To underscore the strong capabilities of our ControlNet Refiner, we present results from various views in Figure \ref{fig:fig9}. Through the integration of ControlNet Refiner, the visual details of the generated objects are significantly enhanced, thereby confirming the efficacy of the ControlNet Refiner for text-to-3D generation.

\paragraph{The Effect of the Consistency Loss. } Lastly, we perform an ablation study on the introduced consistency loss. Specifically, in Figure \ref{fig:fig9}, we showcase the results obtained by incorporating the consistency loss. The example illustrates the effectiveness of this loss function in enhancing consistency across various views of the object. We provide additional experimental results in the Appendix.
\section{Conclusion}
In this paper, we present a novel Gaussian Spatting based text-to-3D approach called DreamPolisher that can generate high-quality and view-consistent 3D assets. In particular, we firstly perform a coarse optimization for initializing the 3D object with the user-provided text prompt. After that, we introduce a ControlNet-based appearance refinement stage and propose a new geometric loss to further improve the texture fidelity and view coherence. Experiments demonstrate that the 3D objects generated by our method show better visual quality and consistency compared with existing methods.
\section{Limitations}
Despite requiring minimal computational resources, DreamPolisher is capable of generating high-fidelity objects with strong coherence across views. However, our approach still exhibits certain limitations: (1) Given that our method uses only 1 GPU for training, DreamPolisher takes about 30 minutes for creating each 3D object. This could be sped up by parallelizing the model; (2) Our method exclusively relies on text instructions as input, without converting them into images as part of the training process. This means it is not possible to guide the generation using images.

\bibliographystyle{splncs04}
\bibliography{main}

\newpage
\appendix 

\section{Appendix}
This Appendix includes the following additional content about:
\begin{enumerate}
\item Further implementation details in Section \ref{implementation}.
\item Additional experimental results in Section \ref{experiment}.
\item Additional visualization results in Section \ref{vis}, as shown in Figure \ref{fig:vis}.
\end{enumerate}
\vspace{-7mm}

\begin{figure}[h!]
    \centering
    \includegraphics[width=1.05\linewidth]{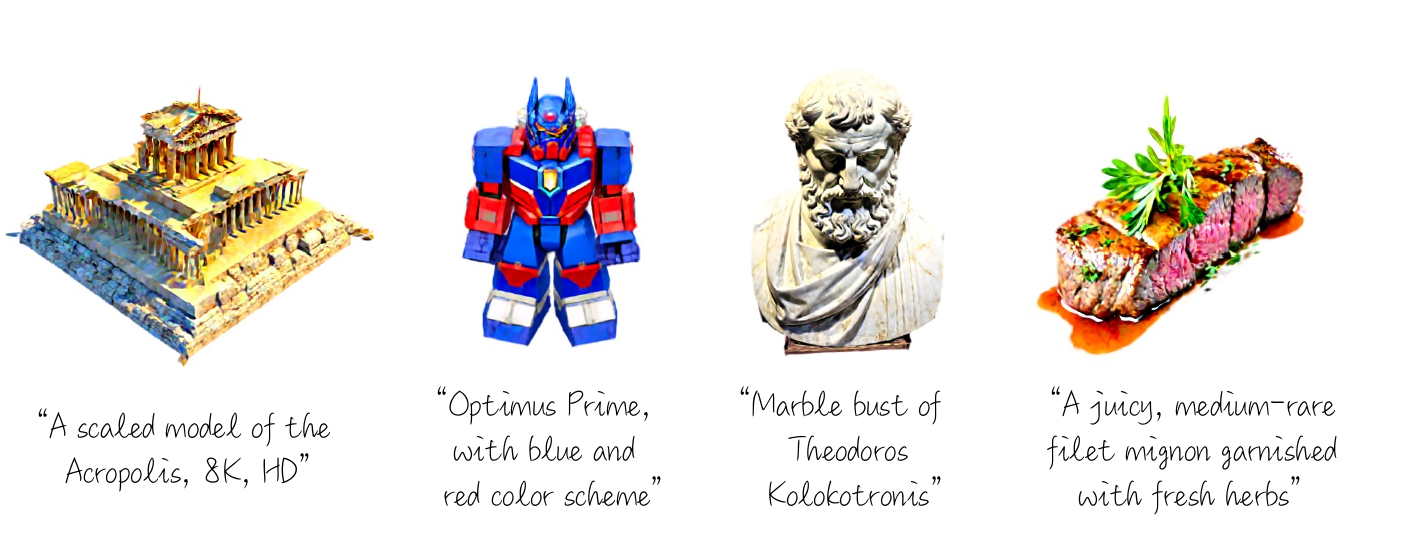}
    \caption{\textbf{Demonstration of the quality and diversity of 3D objects generated by DreamPolisher}. This figure shows high resolution visualizations of the 3D objects generated by our proposed approach, DreamPolisher. For additional views of these examples see Section \ref{experiment}. }
    \label{fig:enter-label}
\end{figure}

\section{Implementation Details}
\label{implementation}
In our main paper's experiments, the training time for each sample is approximately 30 minutes, using a batch size of 4 on 1 $\times$ RTX 3090 GPU. For the diffusion model \cite{rombach2022high}, we set the guidance scale to 7.5. Additionally, we also adopt the negative prompts from \cite{katzir2023noise} following LucidDreamer \cite{liang2023luciddreamer}.

\section{Additional Experimental Results}
\label{experiment}
\subsection{Comparison with existing methods}
\begin{figure*}[t]
\centering
\advance\leftskip-0.3cm
\includegraphics[width=1.03\textwidth]{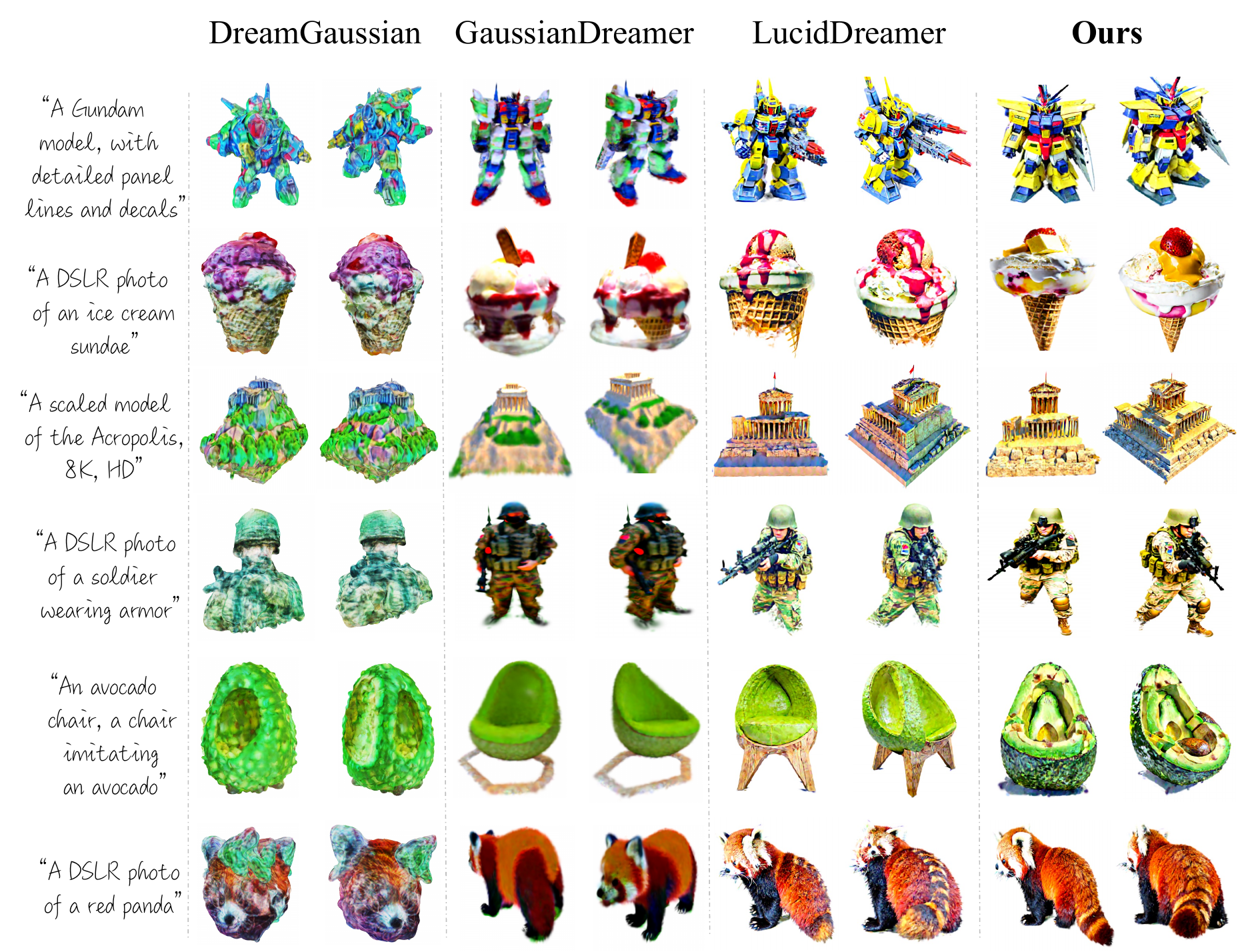}
\vspace{-3mm}
    \caption{\textbf{Comparison with existing methods.} We compare the results generated by DreamPolisher with those from existing methods. To demonstrate the efficiency of DreamPolisher, we include three state-of-the-art approaches for comparison: DreamGaussian \cite{tang2023dreamgaussian}, GaussianDreamer \cite{yi2023gaussiandreamer} and LucidDreamer \cite{liang2023luciddreamer}.}
\vspace{-1mm}
\label{fig:comp}
\end{figure*}

In Figure \ref{fig:comp}, we present additional results comparing our proposed method with state-of-the-art text-to-3D generation baseline approaches, including DreamGaussian \cite{tang2023dreamgaussian}, GaussianDreamer \cite{yi2023gaussiandreamer}, and LucidDreamer \cite{liang2023luciddreamer}.

Our comparison results align with the qualitative findings outlined in the main paper. Specifically, our method consistently produces higher-quality 3D assets that accurately reflect the provided text prompts while maintaining view consistency. On the other hand, the baseline methods exhibit shortcomings in terms of generation quality and consistency when conditioned on the given descriptions. For instance, DreamGaussian and GaussianDreamer often yield low-resolution 3D assets. Although LucidDreamer shows improved quality compared to the former methods, it still grapples with consistency issues, limiting its practical applicability.

\subsection{Additional Ablation Study}
\begin{figure*}[t]
\vspace{10mm}
\centering
\advance\leftskip-0.3cm
\includegraphics[width=1.04\textwidth]{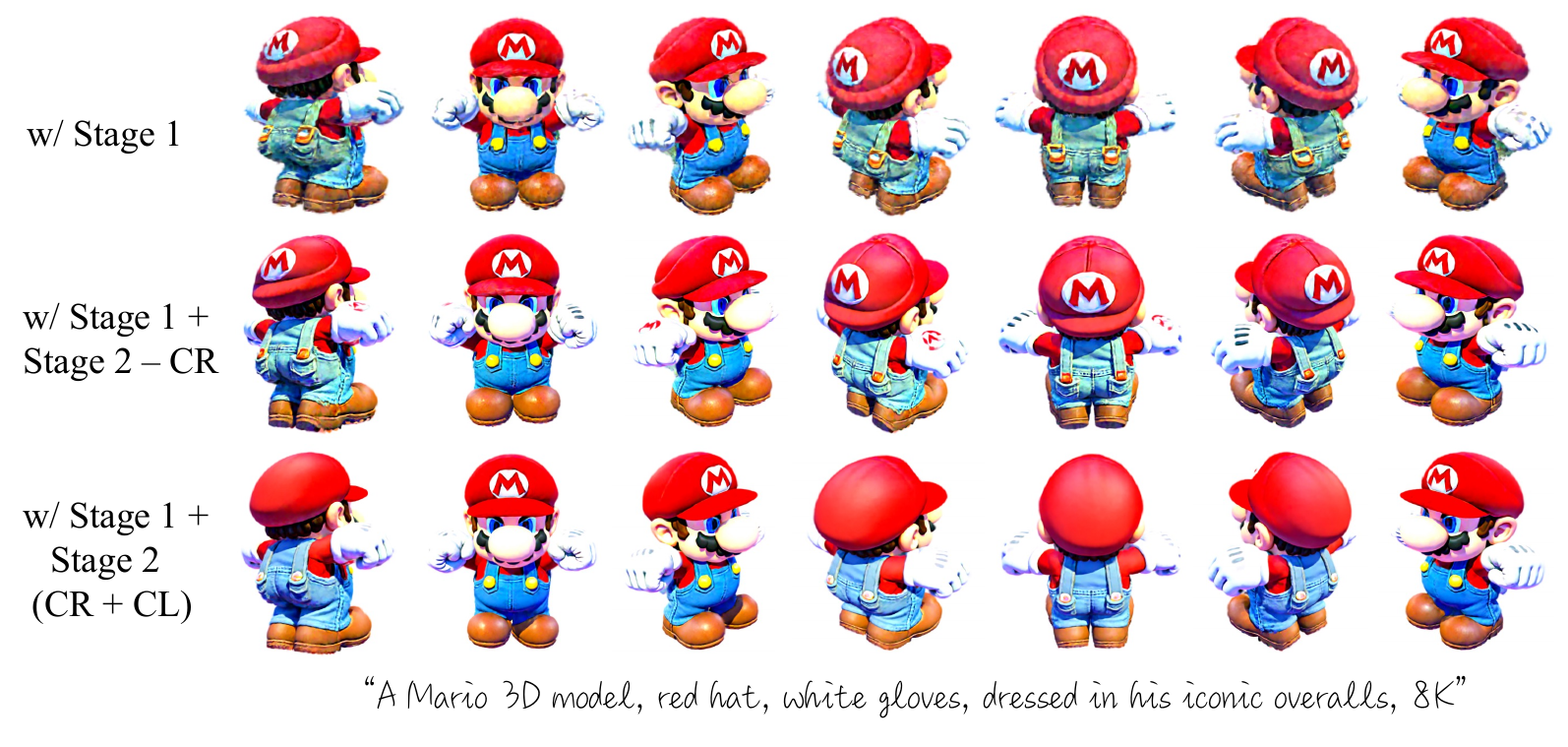}
\vspace{-6mm}
    \caption{\textbf{Ablation study of different components.} We show the results obtained by using various components from our method. Note that ``CR'' and ``CL'' are abbreviations for ControlNet refiner and consistency loss, respectively. ``Stage 2 - CR'' represents only adopting ControlNet refiner in Stage 2, while ``Stage 2 (CR + CL)'' means using ControlNet refiner and consistency loss in Stage 2 together.}
\vspace{-1mm}
\label{fig:sup_ablation}
\end{figure*}

\paragraph{Effect of Different Components. } In Figure \ref{fig:sup_ablation}, we present the results of incorporating various components to validate their significance. It is evident that the ControlNet Refiner notably improves the appearance quality of the 3D assets, while the consistency loss further enhances consistency across different views.

\subsection{Analysis}
\begin{figure*}[t]
\vspace{-50mm}
\centering
\advance\leftskip-0.2cm
\includegraphics[width=1.02\textwidth]{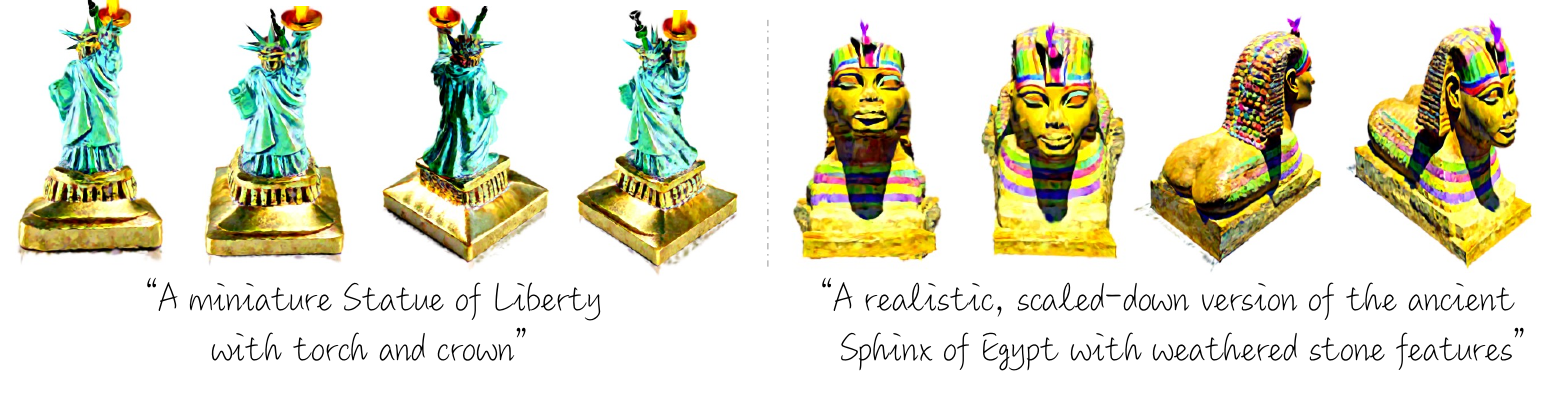}
\vspace{-6mm}
    \caption{\textbf{Failure cases.} Two failure examples generated by the proposed DreamPolisher.}
\vspace{-1mm}
\label{fig:sup_failure}
\end{figure*}

\paragraph{Additional Failure Cases.} 
In Figure~\ref{fig:sup_failure}, we showcase additional failure examples generated by our method. It is apparent that the proposed method may produce unsatisfactory 3D assets when the objects described in the text contain intricate details in localized regions. We hypothesize that this issue arises due to inadequate supervision when solely relying on textual descriptions without incorporating any visual input.

\section{Visualizations}
\label{vis}
\begin{figure*}[t]
\centering
\advance\leftskip-0.2cm
\includegraphics[width=1.04\textwidth]{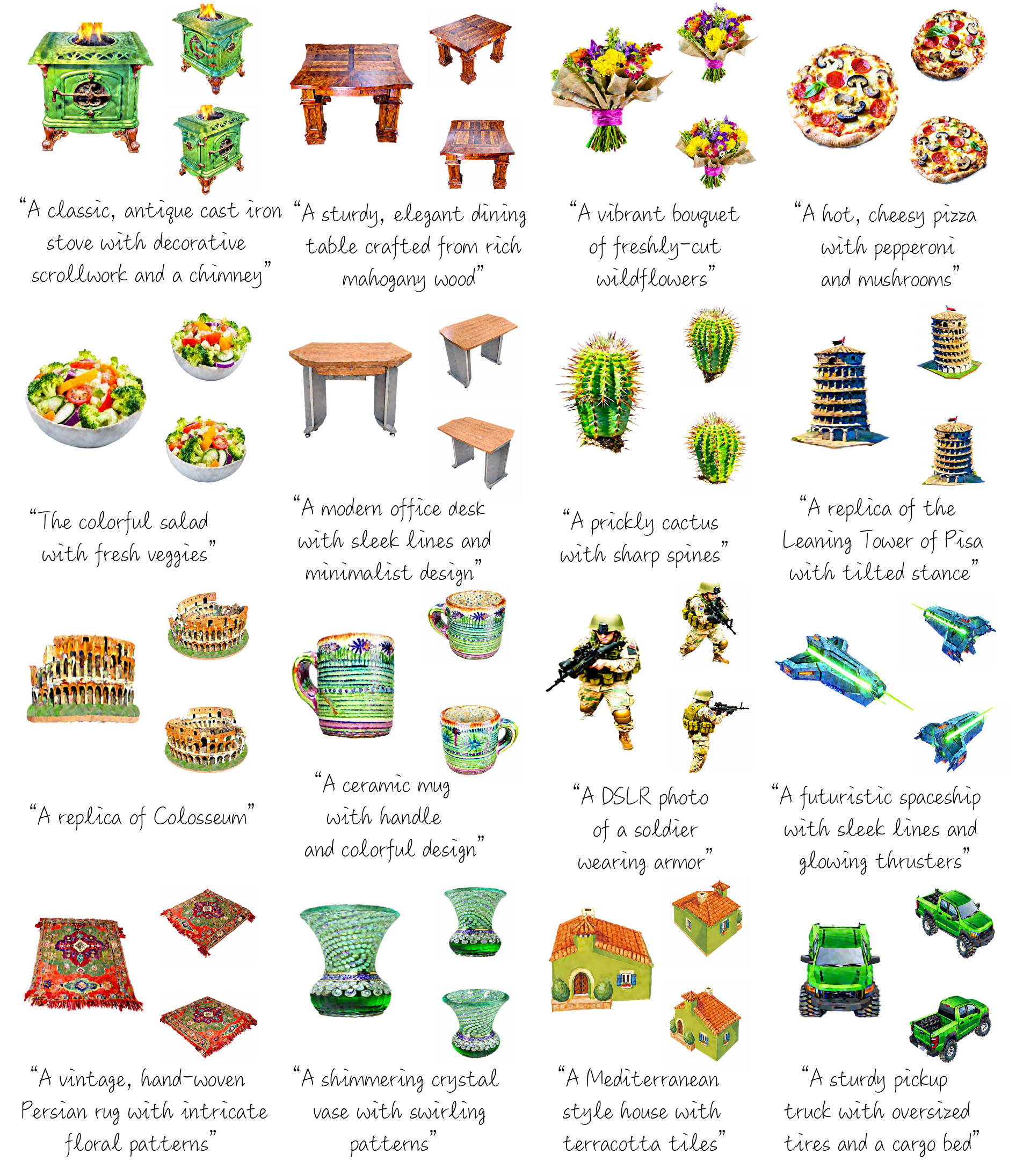}
\vspace{-3mm}
    \caption{\textbf{The generated examples across diverse prompts.} DreamPolisher can produce high-fidelity 3D assets with various prompts, showcasing its strong generality.}
\vspace{-1mm}
\label{fig:vis}
\end{figure*}

We provide additional visualization results of text-to-3D generation from our method in Figure \ref{fig:vis}. Our method benefits from the generality of the introduced geometric diffusion and ControlNet-based refiner, allowing it to generate high-fidelity 3D assets while ensuring consistency across different views. 

\clearpage

\end{document}